\documentclass[runningheads]{llncs}
\usepackage[T1]{fontenc}
\usepackage{graphicx}
\usepackage{booktabs}
\usepackage{amsmath}
\usepackage{amssymb}
\usepackage{amsfonts}
\usepackage{float}
\usepackage{pifont}

\usepackage[misc]{ifsym}
\newcommand{\corr}{(\Letter)}
\usepackage{mwe} 

\usepackage{hyperref} 

\begin{document}

% --- Table of Contents Metadata (Moved here to prevent duplication) ---
\title{Knowledge-Guided Time-Varying Causal Inference for Arctic Sea Ice Dynamics}
\toctitle{Knowledge-Guided Time-Varying Causal Inference for Arctic Sea Ice Dynamics}

% --- Author Section ---
\author{Akila Sampath\inst{1}\corr \and
Vandana P. Janeja\inst{1} \and
Jianwu Wang\inst{1}}

\tocauthor{Akila Sampath, Vandana P. Janeja, Jianwu Wang}
% --- Running Head ---
\authorrunning{A. Sampath et al.}
\titlerunning{Knowledge-Guided Time-Varying Causal Inference}

% --- Affiliations ---
\institute{University of Maryland, Baltimore County, Baltimore, MD, USA \\
\email{\{asampath, vjaneja, jianwu\}@umbc.edu}}

\maketitle            % typeset the header of the contribution
\begin{abstract}
Quantifying the causal relationship between sea ice thickness and sea surface height (SSH) is essential for understanding the mechanisms driving polar climate dynamics. Conventional deep learning models often struggle with treatment effect estimation in climate settings due to time-varying confounding and the lack of physical constraints. To address these challenges, we propose the Knowledge-Guided Causal Model Variational Autoencoder (KGCM-VAE) to quantify the effect of SSH on sea ice thickness. The framework leverages established physical relationships between SSH and surface velocity to generate physically grounded, time-varying continuous treatments, where each treatment value can change at every time step within a sequence. The model also incorporates Maximum Mean Discrepancy (MMD) to balance treated and control distributions in the latent space, mitigating observed confounding bias. Using synthetic data, we evaluated the model's ability to predict sea ice thickness responses under hypothetical SSH forcing scenarios, demonstrating that KGCM-VAE achieves lower PEHE compared to state-of-the-art baselines. Ablation studies further confirm that MMD consistently enhances treatment effect estimation over the base model. Additionally, we conducted a real-world case study to examine the sensitivity of sea ice thickness to SSH forcing and validate our findings against existing physical modeling results.
\end{abstract}

\section{Introduction}
Earth system data exhibit substantial spatial and temporal variability and are generated from both climate model simulations and real-world observations. The increasing volume and complexity of these datasets have driven the adoption of machine learning (ML) and statistical methods to identify patterns, correlations, and complex relationships within Earth system processes \cite{runge2023causal}. However, despite their strong predictive ability, these models primarily uncover surface-level associations, often lacking a theoretical understanding of the cause-and-effect relationships among physical variables \cite{runge2023causal}, which is crucial for process understanding. Deeper understanding requires combining causal inference with predictive machine learning, as it reveals insights that go beyond purely associative patterns, enabling models to generalize more effectively across changing environments and unseen conditions \cite{reichstein2019deep,pearl2018book}. In this context, causal inference provides a framework that combines domain knowledge, machine learning methods, and observational data \cite{pearl2009causality,pearl2018book} to identify the causal structure of a system and estimate the effects of causal relationships. Building on this foundation, we leverage physical domain knowledge to guide causal inference in Arctic climate systems, enabling estimation of causal relationships from real-world observational data.

%%%%This needs to be fixed 
Sea Surface Height (SSH) is a critical indicator of ocean dynamics and sea level rise \cite{kwok2015variability}. Previous studies have identified a strong association between SSH and sea ice thickness \cite{wang2021lasting}; however, the causal nature of this relationship, particularly the mechanisms involving ocean circulation and freshwater distribution, remains poorly understood. Existing approaches have struggled to determine the direction of influence and disentangle the effects of shared atmospheric drivers. Establishing the causal pathway from SSH to sea ice thickness is therefore critical for advancing our understanding of Arctic climate dynamics and improving the decadal predictability of sea ice change. 

To address this challenge, we rely on a qualitative 
literature review \cite{chapman1994arctic,walsh1979interannual,notz2012observations} to inform 
the construction of a preliminary causal time series 
graph that identifies observed and unobserved 
confounding variables and potential connections 
between Arctic climate drivers and sea ice thickness. 
Building on this graph, we propose the Knowledge-Guided 
Causal Modeling Variational Autoencoder (KGCM-VAE), 
a VAE-based architecture designed to provide accurate 
estimates of the total causal effect in such complex, 
indirect systems.

\paragraph{\textbf{Motivation}}
Our work is motivated by three key considerations.
\textbf{(1) Limitations of linear and static causal models.}
Conventional causal methods assume linear, time-invariant dynamics~\cite{peters2017elements,pearl2018book} that are poorly suited for the nonlinear and seasonally evolving Arctic ocean–ice system~\cite{chapman1994arctic,walsh1979interannual}. Data-driven causal frameworks can instead capture heterogeneous, time-varying treatment effects without strong parametric assumptions~\cite{shalit2017estimating,louizos2017causal}.
\textbf{(2) Limited support for continuous, lagged treatments.}
Most temporal causal frameworks assume discrete treatments, whereas SSH forcing is continuous and induces delayed effects on sea ice thickness. Few methods jointly model continuous treatments, temporal lags, and ITEs in sequential settings. KGCM-VAE addresses this by conditioning on the full treatment history $\mathbf{T}_{t-\text{Lag}+1:t}$ to capture delayed causal responses based on physical formulation.
\textbf{(3) Limitations of propensity score methods.}
Propensity score approaches~\cite{rosenbaum1984association} are tailored to binary treatments and scale poorly to continuous, time-varying settings, requiring high-dimensional density estimation at each timestep while failing to leverage temporal structure. Instead, we apply Maximum Mean Discrepancy (MMD) regularization in latent space to balance representations without explicit propensity estimation, enabling efficient learning over long sequences.
\paragraph{\textbf{Main Contributions}}
First, we introduce KGCM-VAE, a physics-guided VAE framework for causal representation learning in sequential climate systems. Second, unlike traditional models that focus primarily on input reconstruction, our approach integrates MMD based latent deconfounding within a recurrent VAE architecture, enabling estimation of time-varying causal effects. Third, the framework incorporates established physical relationships, such as those between SSH and ice velocity, to guide treatment generation and ensure physically consistent counterfactual predictions. We demonstrate that KGCM-VAE achieves lower PEHE than state-of-the-art baselines on both synthetic and real-world Arctic data.

\section{Related Work}

Climate research often relies on physics-based simulation 
models to investigate causal mechanisms and system dynamics, 
but the reliability of these findings depends on how 
accurately the simulated physics represents real-world 
processes. Consequently, there is a critical demand for 
deriving causal connections directly from real-world 
time series 
\cite{runge2019detecting,runge2023causal,wu2022nonlinear}. 
However, existing data-driven approaches remain largely 
disconnected from the physical principles governing 
Earth system processes, creating a gap between 
statistical causal discovery and physically grounded 
causal modeling.

In Earth and climate sciences, causal inference 
has been applied primarily for causal discovery, 
which seeks to identify the structure of causal graphs 
from observational time series 
\cite{runge2019detecting,runge2023causal,wu2022nonlinear}. 
These approaches focus on determining which variables 
influence others rather than quantifying the magnitude 
of causal effects under intervention. Furthermore, 
existing Earth system causal methods largely assume 
linear or discrete treatment settings, limiting their 
applicability to problems requiring estimation of 
continuous and lagged treatment effects informed by 
physical domain knowledge. More broadly, treatment 
effect estimation methods can be categorized into 
parametric and non-parametric approaches.

In non-parametric approaches such as nearest-neighbor 
matching, propensity score matching, and propensity 
score re-weighting, the context-intervention-outcome 
relationship is not modeled explicitly 
\cite{austin2011introduction}. Parametric methods such 
as linear and logistic regression, as well as structured 
models like random forests and regression trees, build 
explicit mappings to model the relationship between 
covariates, treatment, and outcomes 
\cite{chipman2010bart,wager2018estimation}. However, 
weighting-based methods can exhibit high variance in 
effect estimates under poor covariate overlap between 
treatment groups, as propensity scores near zero or 
one lead to high inverse probability weights. Recent 
advances have increasingly leveraged machine learning 
techniques to improve treatment effect estimation in 
high-dimensional settings.

Tree-based machine learning methods are increasingly 
utilized for causal inference, particularly in the 
estimation of individual-level treatment effects 
\cite{wager2018estimation,athey2016recursive}. 
Furthermore, researchers have shown that high 
dimensional regression techniques like Lasso can be 
adapted to reliably estimate treatment effects, 
attaining convergence rates comparable to those in 
semi-parametric settings 
\cite{belloni2014inference,athey2016recursive}. 
Representation learning offers a compelling approach 
to counterfactual inference by learning embeddings 
that help address the challenge of unobserved 
potential outcomes \cite{johansson2016learning}.

Although the SITE (Similarity-Preserved Individual 
Treatment Effect) framework leverages deep 
representation learning for estimating individual 
treatment effects, it generally lacks mechanisms to 
capture intricate time-lagged temporal relationships 
\cite{yao2018representation}. Existing 
representation-learning approaches often struggle 
to preserve physically meaningful similarity 
information distributed across space and time, 
particularly when interactions are highly nonlinear. 
Although \cite{tian2014simple} demonstrated that 
balance can be achieved through strategic covariate 
selection and by modeling the interactions between 
treatments and covariates, many existing frameworks 
fail to jointly model time-varying treatment effects 
and domain-specific physical constraints.

To address these gaps, our proposed method integrates 
physical domain knowledge directly into the causal 
modeling framework, enabling treatment effect 
estimation that remains consistent with established 
Arctic ocean-ice processes while capturing continuous, 
time-varying causal dynamics and seasonal 
non-stationarity.

\section{Methodology}

\subsection{Problem Formulation}

The primary objective of this study is to quantify the causal effect of short-term variations in SSH (treatment $T$) on Arctic sea ice thickness (outcome $Y$) using a knowledge-guided causal modeling framework that accounts for historical temporal dependencies. Given a historical look-back window of length $\text{Lag}$, we define the input space at time step $t$ as a sequence of historical covariates $\mathbf{X}_{t-\text{Lag}+1:t}$ and historical treatments $\mathbf{T}_{t-\text{Lag}+1:t}$. The covariates comprise oceanic feature variables such as sea surface velocities and raw SSH observed over the past $\text{Lag}$ time steps. We distinguish between two treatment sequences: the factual (observed) unmodulated signal $T_0$, and the counterfactual (intervened) velocity-modulated signal $T_1$, which simulates a scenario in which oceanic transport is systematically enhanced, representing a physically motivated causal intervention.

To capture temporal dynamics, the potential outcomes are defined as a function of the historical trajectories of both covariates and treatments:
\begin{equation}
\hat{Y}_{t,\text{Lag}}(T_j) = f\!\left(\mathbf{X}_{t-\text{Lag}+1:t},\, \mathbf{T}_{j,\,t-\text{Lag}+1:t}\right),
\end{equation}
where $\hat{Y}_{t,\text{Lag}}(T_j)$ is the predicted potential outcome for treatment $T_j$, estimated at time $t$ over a historical window of length $\text{Lag}$, and $j \in \{0, 1\}$ indexes the factual and counterfactual treatment sequences, respectively.
 The Individual Treatment Effect (ITE) is then estimated at each time step $t$ by contrasting the two potential outcomes:
\begin{equation}
\tau_{t} = \hat{Y}_{t,\text{Lag}}(T_1) - \hat{Y}_{t,\text{Lag}}(T_0),
\end{equation}
\subsection{Causal Assumptions}
\begin{figure}[htbp]
    \centering
    % First Image
    \includegraphics[width=0.75\textwidth]{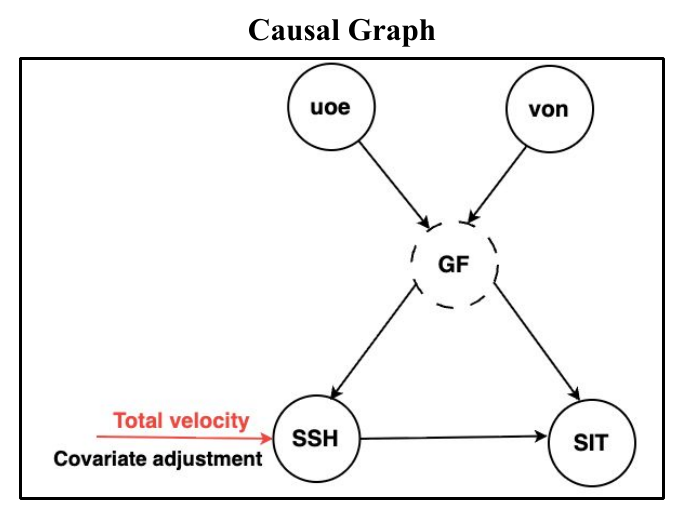}
   
\caption{The underlying causal graph with physical variables explaining the physical mechanism, with geostrophic flow (GF) as the unmeasured confounder.}
    \label{fig:causalgraph}
\end{figure}

The following assumptions underpin identification of the 
causal effect of SSH on sea ice thickness, as represented 
in the causal graph (Figure~\ref{fig:causalgraph}).
\paragraph{\textbf{Causal Directed Acyclic Graph (DAG).}}
The relationships are represented by a DAG without feedback loops. Atmospheric forcing precedes both SSH and sea ice thickness, and total velocity conditions the SSH treatment. Time flows strictly forward, enforced through the treatment lag in data construction.

\paragraph{\textbf{No Unmeasured Confounding.}}
Conditional on $\text{total velocity}$, no remaining common causes of SSH and sea ice thickness are assumed to exist:
\begin{equation}
    Y(t) \perp SSH \mid \text{total velocity}
\end{equation}
Total velocity serves as a sufficient proxy for the atmospheric circulation state driving SSH anomalies via mass redistribution and sea ice thickness via heat advection.

\paragraph{\textbf{Positivity (Overlap).}}
Each unit has non-zero probability of receiving any level of SSH treatment conditional on total velocity:
\begin{equation}
    0 < p(SSH = t \mid \text{total velocity}) < 1, \quad \forall\, t \in \mathcal{T}
\end{equation}
This is enforced computationally using MMD regularization, which promotes overlap between treated and control latent representations.
\paragraph{\textbf{Consistency (SUTVA).}}
The potential outcome for sea ice thickness at time $t$ is determined solely by the unit's own SSH treatment trajectory and $\text{total velocity}$ history, with no interference across units or time steps. Each treatment level corresponds to a unique, well-defined SSH regime.
The causal structure adopted in this work is grounded 
in established Arctic ocean-ice physics rather than 
purely statistical assumptions. Specifically, it embeds 
the hydrostatic equilibrium governing SSH and sea ice 
thickness, alongside the geostrophic balance linking 
SSH gradients to surface velocity. By integrating these 
physical relationships directly into the model architecture 
(Section~\ref{sec:knowledge_guidance}), the causal 
structure reflects mechanistically plausible pathways 
derived from domain theory.
\subsection{Physical Knowledge Integration}
\label{sec:knowledge_guidance}
The KGCM-VAE framework incorporates domain-specific physical constraints to guide the learning of causal relationships within the Arctic system. For this purpose, we adapt the guidance based on the discussion~\cite{kwok2015variability} and the established relationship between SSH and sea ice thickness, which is essential for ensuring the physical consistency of causal estimates in ice-covered oceans. This interaction is governed by the following hydrostatic equilibrium equation:
\begin{equation}
h_i = \left(f_{\text{b}} - h_{\text{SSH}}\right) \left(\frac{\rho_w}{\rho_w - \rho_i}\right) - h_s \left(\frac{\rho_w - \rho_s}{\rho_w - \rho_i}\right),
\end{equation}
where $h_i$ denotes sea ice thickness, $f_{\text{b}}$ represents ice freeboard (ice above the waterline), $h_{\text{SSH}}$ is the sea surface height, and $h_s$ denotes snow depth. The parameters $\rho_w$, $\rho_i$, and $\rho_s$ correspond to the densities of seawater, sea ice, and snow, respectively.

The framework further leverages the connection between SSH and surface velocity~\cite{doglioni2023sea} to capture the underlying Arctic ocean dynamics. This relationship is formalized through geostrophic balance, where the surface velocity components ($u_g$, $v_g$) are determined by the spatial gradients of SSH:
\begin{equation}
u_g = -\frac{g}{f_c}\frac{\partial \eta}{\partial y}, \qquad v_g = \frac{g}{f_c}\frac{\partial \eta}{\partial x},
\end{equation}
where $g$ is the gravitational acceleration, $f_c$ is the Coriolis parameter, and $\partial\eta/\partial x$ and $\partial\eta/\partial y$ represent the spatial gradients of SSH. The total velocity field acts as an adjustment variable that conditions the SSH treatment and blocks the backdoor path induced by the unmeasured confounder, geostrophic flow, which influences both SSH and sea ice thickness. This is formalized in the causal graph shown in Figure~\ref{fig:causalgraph}, where changes in SSH are assumed to causally drive changes in sea ice thickness, geostrophic flow represents the unmeasured confounder, and total velocity serves as its observed proxy for adjustment. Accordingly, we assume that total velocity functions as a confounding variable influencing both the SSH and sea ice thickness fields. Our causal graph further supports the assumption that changes in SSH influence changes in sea ice thickness while accounting for pressure gradient as an unmeasured confounder.

\subsection{Knowledge-Guided Time-Varying Treatment Generation}
To capture sub-seasonal variability and reduce high-frequency noise, a moving window filter is first applied to the raw SSH signal, yielding a smoothed $\widetilde{\mathrm{SSH}}_t$. This smoothed signal is then dynamically amplified based on ocean circulation intensity, represented by the total surface velocity magnitude. The underlying physical mechanism assumes that stronger surface currents drive SSH gradients, subsequently amplifying the forcing exerted on the sea ice. To model this non-linear, threshold-dependent response, we employ a sigmoid function, which provides a smooth, bounded, and differentiable transition between low and high amplification regimes.
The amplification is governed by the modulation factor $\gamma_t$, computed as:
\begin{equation}
\gamma_t = \frac{1}{1 + \exp\!\left[-a\,(\tilde{v}_t - v_0)\right]},
\label{eq:sigmoid_modulation}
\end{equation}
where $\tilde{v}_t$ is the smoothed total velocity, $v_0$ is the mean velocity for amplification, and $a$ is the steepness parameter controlling the sharpness of the sigmoid response.

The smoothed SSH signal is then modulated by the factor $(1 + \beta\gamma_t)$, where $\beta$ is the overall modulation strength parameter. The resulting velocity-modulated treatment signal ($\mathrm{SSH}_t^{\text{treat}}$) is given by:
\begin{equation}
\mathrm{SSH}_t^{\text{treat}} = (1 + \beta\gamma_t)\,\widetilde{\mathrm{SSH}}_t,
\label{eq:final_treated_signal}
\end{equation}
The proposed treatment is defined as a non-linear function of the total velocity ($\tilde{v}_t$), capturing the inherent state-dependency of climate impacts. In accordance with established fluid dynamics, incremental variations in SSH may yield negligible dynamical consequences at low velocities; however, these effects can become catastrophic once the system surpasses a critical threshold ($v_0$). This formulation is physically consistent with the tipping-point dynamics observed in ocean-ice interactions where small perturbations near a regime shift can trigger disproportionately large responses. 
\begin{figure}[hbt!]
\centering
\includegraphics[width=1.07\textwidth]{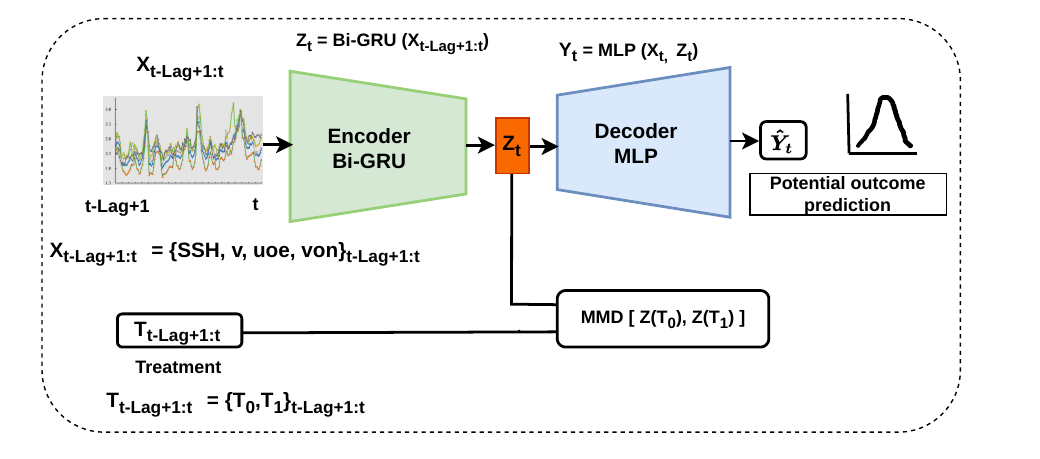}
\caption{Overview of the proposed Knowledge-Guided Causal Model Variational Autoencoder (KGCM-VAE). The covariates are Eastward velocity (uoe), Northward velocity (von), Total velocity (v), Sea Surface Height (SSH), physics-guided treatment ($T = \{T_0, T_1\}$), and potential outcome ($\hat{Y}$).}
\label{fig:archi}
\end{figure}
\subsection{Model Architecture}

\paragraph{\textbf{Overview.}}
The Knowledge-Guided Causal Modeling VAE (KGCM-VAE) (Figure~\ref{fig:archi}) is a temporal causal inference model designed to estimate individualized treatment effects from time-varying observational data. The name reflects two core design principles: knowledge-guided, referring to the use of known Arctic physical formulations to construct the causal treatment generation mechanism (SSH forcing modulated by ice velocity via a sigmoid function); and causal modeling, referring to the VAE-based latent space regularization that enables counterfactual inference. The model extends the standard Variational Autoencoder (VAE) framework~\cite{louizos2017causal} by conditioning both the encoder and decoder on a continuous time-varying treatment signal, and by imposing MMD regularization in the latent space to achieve treatment balance. The architecture processes sequences of length $T = 30$ timesteps and produces potential outcome predictions at every timestep, enabling temporally-resolved causal effect estimation throughout the full sequence.

\paragraph{\textbf{Input Representation.}}
The model receives two inputs at each timestep $t$: the covariate vector $\mathbf{X}_t \in \mathbb{R}^{p}$ and the continuous treatment value $T_t \in \mathbb{R}$. The covariate vector comprises $p = 6$ features: eastward ocean velocity ($u$), northward ocean velocity ($v$), total surface velocity, SSH, factual treatment $T_0$, and its lagged value $T_{0,\text{Lag}}$. Importantly, the treatment $T_0$ is included within the covariate vector $\mathbf{X}_t$, and is additionally passed as a separate explicit input to both the encoder and the decoder. This dual inclusion ensures the model captures treatment dynamics both in the temporal context and in the outcome prediction stage.

\paragraph{\textbf{Encoder: $q(\mathbf{Z} \mid \mathbf{X}, T)$.}}
The encoder maps the input sequence to a distribution over latent variables at every timestep. At each timestep $t$, the covariate vector $\mathbf{X}_t$ and the treatment value $T_t$ are concatenated to form a joint input:
\begin{equation}
    \tilde{\mathbf{x}}_t = [\mathbf{X}_t \,;\, T_t] \in \mathbb{R}^{p+1},
\end{equation}
This concatenated sequence is processed by a Bidirectional Gated Recurrent Unit (Bi-GRU) encoder with hidden dimension $d_h = 128$, capturing both forward and backward temporal dependencies across the full sequence:
\begin{equation}
    \mathbf{H} = \text{Bi-GRU}\!\left(\tilde{\mathbf{x}}_{1:T}\right), \qquad \mathbf{H} \in \mathbb{R}^{B \times T \times 2d_h}
\end{equation}
To capture the system's trajectory throughout the entire sequence, the Bi-GRU produces a hidden state at every timestep. This dense temporal information, represented by the sequence $\mathbf{H}$, is then passed through dual linear projection layers to define the parameters of the approximate posterior:
\begin{equation}
    \boldsymbol{\mu}_t = f_{\mu}(\mathbf{H}_t), \qquad \log\boldsymbol{\sigma}^2_t = f_{\sigma}(\mathbf{H}_t),
\end{equation}
where $f_{\mu}$ and $f_{\sigma}$ are single linear layers mapping $2d_h \rightarrow d_z$, outputting the mean and log-variance of the variational posterior $q(\mathbf{Z} \mid \mathbf{X}, T)$ at each timestep. The latent vector is sampled via the reparameterization trick:
\begin{equation}
    \mathbf{Z}_t = \boldsymbol{\mu}_t + \boldsymbol{\epsilon} \odot \exp\!\left(0.5\,\log\boldsymbol{\sigma}^2_t\right), \qquad \boldsymbol{\epsilon} \sim \mathcal{N}(0, \mathbf{I})
\end{equation}
The latent dimension is set to $d_z = 64$. Because the encoder receives both $\mathbf{X}_t$ and $T_t$, the latent representation $\mathbf{Z}_t$ captures treatment-aware environmental states, which is a prerequisite for MMD-based treatment balancing.

\paragraph{\textbf{Decoder: $p(Y \mid \mathbf{Z}, T)$.}}
The decoder predicts the outcome at every timestep by conditioning on both the latent state $\mathbf{Z}_t$ and the treatment value $T_t$. The treatment is first projected from its raw scalar value to a 16-dimensional embedding via a learned treatment projection layer:
\begin{equation}
    \mathbf{t}_{\text{emb},t} = \text{ReLU}\!\left(\mathbf{W}_{\text{proj}}\,T_t + \mathbf{b}_{\text{proj}}\right), \qquad \mathbf{t}_{\text{emb},t} \in \mathbb{R}^{16}
\end{equation}
This projection amplifies the treatment signal before concatenation with the 64-dimensional latent vector, preventing the treatment from being overwhelmed by the higher-dimensional latent representation. The projected treatment and latent vector are concatenated and passed through a three-layer MLP outcome head:
\begin{equation}
    \hat{Y}_t = \text{MLP}\!\left([\mathbf{Z}_t \,;\, \mathbf{t}_{\text{emb},t}]\right),
\end{equation}
with architecture $(d_z + 16) \rightarrow 128 \rightarrow 64 \rightarrow 1$, using ReLU activations and dropout ($p = 0.1$) after the first layer. This single outcome head conditioned on treatment enables counterfactual prediction by substituting the treatment value at inference time:
\begin{equation}
\hat{Y}_{0,t} = \text{MLP}\Big([\mathbf{Z}_t \,;\, \mathbf{t}_{\text{proj}}(T_{0,t})]\Big), \quad
\hat{Y}_{1,t} = \text{MLP}\Big([\mathbf{Z}_t \,;\, \mathbf{t}_{\text{proj}}(T_{1,t})]\Big),
\end{equation}
The Individual Treatment Effect (ITE) at each timestep is then estimated as:
\begin{equation}
    \widehat{\text{ITE}}_t = \hat{Y}_{1,t} - \hat{Y}_{0,t},
\end{equation}
This design enables temporally-resolved causal effect estimation across all $T = 30$ timesteps in a sequence.

\paragraph{\textbf{MMD Deconfounding Regularization.}}
A core challenge in observational causal inference is that treatment assignment is not random. In the Arctic system, SSH variations (treatment) are inherently correlated with surface velocity (confounder), which independently influences sea ice thickness (outcome). To address this, MMD regularization is imposed on the latent space $\mathbf{Z}$ to enforce treatment balance. The continuous treatment $T_t$ is thresholded at its batch median to define control and treated latent groups:
\begin{equation}
    \mathbf{Z}_0 = \{\mathbf{z}_t \mid T_t \leq \text{median}(T)\}, \qquad \mathbf{Z}_1 = \{\mathbf{z}_t \mid T_t > \text{median}(T)\},
\end{equation}
The empirical MMD$^2$ between the two groups is computed using a multi-scale RBF kernel with bandwidths $\lambda \in \{0.5, 1.0, 2.0\}$:
\begin{equation}
    \mathcal{L}_{\text{MMD}} = \frac{1}{3}\sum_{\lambda \in \{0.5,\,1.0,\,2.0\}} \text{MMD}^2(\mathbf{Z}_0, \mathbf{Z}_1;\,\lambda),
\end{equation}
\begin{equation}
    \text{MMD}^2(\mathbf{Z}_0,\mathbf{Z}_1) = \frac{1}{n(n-1)}\sum_{i \neq j}k(z^0_i,z^0_j) + \frac{1}{m(m-1)}\sum_{i \neq j}k(z^1_i,z^1_j) - \frac{2}{nm}\sum_{i,j}k(z^0_i,z^1_j),
\end{equation}
where $k(x,y) = \exp\!\left(-\|x-y\|^2 / 2\lambda^2\right)$ is the RBF kernel and $n$, $m$ are the sizes of the control and treated groups within each mini-batch. Minimizing $\mathcal{L}_{\text{MMD}}$ compels the encoder to produce latent representations that are statistically indistinguishable across treatment levels, enabling reliable counterfactual prediction.

\paragraph{\textbf{KL Divergence Regularization.}}
The variational posterior $q(\mathbf{Z} \mid \mathbf{X}, T)$ is regularized toward a standard Gaussian prior $p(\mathbf{Z}) = \mathcal{N}(0, \mathbf{I})$ via the KL divergence:
\begin{equation}
    \mathcal{L}_{\text{KL}} = -\frac{1}{2}\sum_{j=1}^{d_z}\left(1 + \log\sigma_j^2 - \mu_j^2 - \sigma_j^2\right),
\end{equation}
This regularization prevents the encoder from collapsing to a deterministic mapping and encourages a smooth, continuous latent space in which nearby points correspond to similar physical states.

\paragraph{\textbf{Training Objective.}}
The KGCM-VAE is trained end-to-end by minimizing the composite loss
\begin{equation}
\mathcal{L}_{\text{total}} 
= \beta_{\text{rec}} \mathcal{L}_{\text{MSE}} 
+ \beta_{\text{KL}} \mathcal{L}_{\text{KL}} 
+ \beta_{\text{MMD}} \mathcal{L}_{\text{MMD}},
\end{equation}
The reconstruction term $\mathcal{L}_{\text{MSE}}$ = $\frac{1}{N}\sum_{i=1}^{N}(\hat{Y}_i - Y_i)^2$. We set $\beta{\text{rec}} = 10$, $\beta_{\text{KL}} = 0.001$, and $\beta_{\text{MMD}} = 1.0$. To stabilize training, $\beta_{\text{KL}}$ and $\beta_{\text{MMD}}$ follow linear warmup schedules over 30 and 50 epochs, respectively. The KL divergence term is restricted to a maximum value of 1.0 to prevent the latent representations from collapsing to the prior distribution.
\section{Experiments}

\subsection{Evaluation Metrics}
We evaluate KGCM-VAE using Precision in Estimation 
of Heterogeneous Effects ($PEHE_\text{Lag}$) for 
causal effect estimation and Root Mean Squared Error 
(RMSE) for overall predictive accuracy. Since ground 
truth counterfactual outcomes are unobservable in 
real-world settings, $PEHE_\text{Lag}$ is evaluated 
on synthetic data where both potential outcomes 
$Y^{(1)}$ and $Y^{(0)}$ are known by construction:
\begin{equation}
PEHE_\text{Lag} = \frac{1}{N} \sum_{i=1}^{N} 
\left( \hat{\tau}_i(t, \text{Lag}) - 
\tau_i(t, \text{Lag}) \right)^2,
\end{equation}
where $\hat{\tau}_i(t, \text{Lag})$ is the predicted 
ITE and $\tau_i(t, \text{Lag})$ is the ground truth 
ITE. The ground-truth lagged ITE 
at prediction horizon Lag $\text{Lag}$ is defined as
\begin{equation}
\tau(t, \text{Lag}) = Y^{(1)}(t + \text{Lag} \mid 
A_t =1, X_{0:t}) - Y^{(0)}(t + \text{Lag} \mid 
A_t = 0, X_{0:t}),
\end{equation}
where $Y^{(1)}$ and $Y^{(0)}$ denote the potential 
outcomes under treatment and control conditions 
respectively, and $t + \text{Lag}$ is the future 
time step for the sea ice thickness outcome. This 
metric penalizes discrepancies between predicted 
and ground truth ITEs, capturing the delayed sea 
ice response to historical treatments.
\subsection{Baselines}
To evaluate the performance of KGCM-VAE, we compare it against three representative deep learning architectures for time-series causal inference that are capable of estimating counterfactual outcomes for ITE analysis. CRN~\cite{bica2020estimating} learns disentangled temporal representations that are predictive of outcomes while remaining invariant to treatment assignment. Following the causal RNN architecture proposed in \cite{poulos2021rnn}, this baseline uses a sequence-to-sequence RNN to estimate counterfactual outcomes. The model employs an encoder to compress the observed history into a fixed-dimensional context vector, which then initializes a decoder to predict potential outcomes under alternative treatment sequences. G‑Net~\cite{li2021g} extends RNNs to implement g‑computation for dynamic treatment regimes and estimates expected counterfactual outcomes under time‑varying treatment strategies. For a fair comparison, we adopted the LSTM based architecture used in this study.

To ensure a fair comparison, we adapt all baselines to handle time-varying confounding. Code is available \href{https://github.com/akilasampath5/KGCM-VAE}{here}.
\subsection{Data}
The primary data source for this research is the Subseasonal to Seasonal (S2S) ECMWF Real-time Control Forecast \cite{peng2023skill}. This dataset provides a critical basis for describing short-term variations in ocean dynamics. The study focuses on a temporal span of 1,620 days, covering the period from January 2020 through June 2024. To isolate the physical dynamics of the Arctic environment, the global 1° grid ($360 \times 181$) was subsetted to the high northern latitudes, specifically the region between 60°N and 90°N. To address the computational complexity and focus on regional-scale physical trends, a spatial averaging preprocessing step was implemented. This involved computing daily time series by averaging each physical parameter including sea-ice thickness, SSH, and surface seawater velocities (eastward, northward, and total) across the spatial grid. 

\subsection{Synthetic Ground Truth Data Generation}

The true causal effect is unobservable in real-world 
Arctic datasets. Consequently, we construct a 
physics-guided synthetic data formulation in which 
the treatment effect is explicitly specified, enabling 
direct evaluation of causal effect estimation while 
maintaining consistency with known Arctic sea ice 
processes. The time-varying causal effect is 

\begin{equation}
\delta_t = -\alpha |\eta_{t}| \tanh\!\big(2(T_{1,t} - T_{0,t})\big),
\end{equation}

where $\alpha$ is a hyperparameter controlling the effect magnitude, the nonlinear dependence on the treatment contrast $T_{1,t} - T_{0,t}$ reflects realistic ocean-ice interaction dynamics, the unmeasured seasonal confounder is modeled as $\eta_{t} = \sin\!\left(\frac{30\pi t}{N}\right)$, and the counterfactual outcome under treatment is generated as $Y_{1,t} = Y_{0,t} + \delta_t$.
\begin{table}[ht]
\centering
\caption{\textbf{Performance Comparison Against State-of-the-Art Baselines}}
\label{tab:benchmark_comparison}
\small
\setlength{\tabcolsep}{3pt}
\renewcommand{\arraystretch}{0.65}
\begin{tabular}{@{}l c|c c|c c|c@{}}
\toprule
& \multicolumn{2}{c}{\textbf{\text{Lag} = 3}} 
& \multicolumn{2}{c}{\textbf{\text{Lag} = 6}} 
& \multicolumn{2}{c}{\textbf{\text{Lag} = 9}} \\
\cmidrule(lr){2-3} \cmidrule(lr){4-5} \cmidrule(lr){6-7}
\textbf{Model} 
& \textbf{RMSE} & \textbf{PEHE} 
& \textbf{RMSE} & \textbf{PEHE} 
& \textbf{RMSE} & \textbf{PEHE} \\
\midrule
CRN     & 0.9814 & 3.1353 &  0.9792 & 3.1583  & 0.9790 & 3.1522 \\
Causal-RNN   &  0.9933 & 3.2119  & 0.9928 & 3.2136 & 0.9931 & 3.2155  \\
G-Net &  0.9994 & 3.1660  & 1.0004 & 3.1683  &  1.0017 & 3.1679\\
\midrule
\textbf{KGCM-VAE (Ours)} 
& {0.9883 } & \textbf{3.0804}  
& {0.9884 } & \textbf{3.0816 } 
& {0.9880 } & \textbf{3.0840 } \\
\bottomrule
\end{tabular}
\end{table}

This formulation generates synthetic data where SSH anomalies are sinusoidal in nature \cite{gill2016atmosphere}, ensuring the incorporation of physical dynamics. By modeling $\eta_{t}$ as a periodic variation, we ensure that our causal framework respects the fundamental dynamics of the sea-ice system prior to applying the model to real-world sea ice observations. Because the true causal effect $\delta_t$ is known by construction, this setup enables precise evaluation of counterfactual prediction accuracy and PEHE, integrating physics-guided modeling with controlled 
causal validation.
\begin{figure}[ht!]
    \centering
    % First Image
    \includegraphics[width=0.75\textwidth]{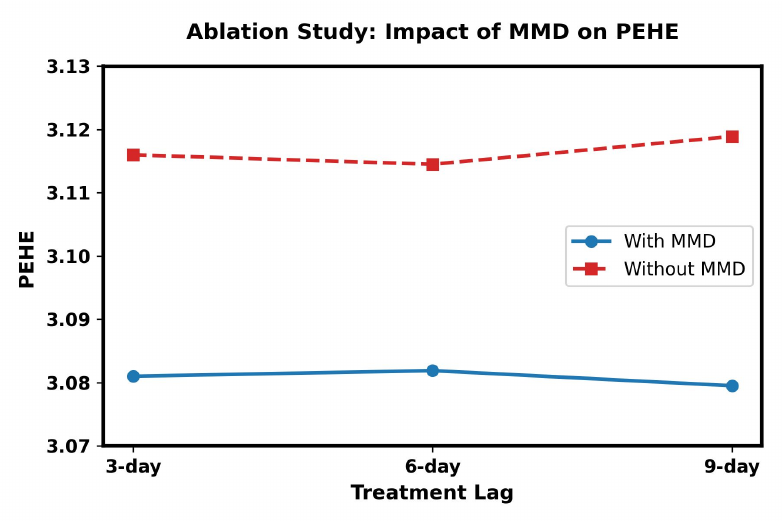}
   
\caption{Impact of MMD regularisation on PEHE across treatment lags, comparing model performance with and without MMD.}
    \label{fig:ablation_unified}
\end{figure}
\subsection{Results and Evaluation on Synthetic Data}

As illustrated in Table \ref{tab:benchmark_comparison}, 
KGCM-VAE consistently outperforms state-of-the-art baselines 
(CRN, Causal-RNN, and G-Net) in counterfactual estimation, 
achieving the lowest PEHE across all treatment lags. Although 
the absolute differences appear modest, this is expected given 
that spatially averaged time series data inherently smooths 
out local variability, compressing the range of counterfactual 
responses and narrowing the margin between methods. The consistent ranking of KGCM-VAE across all lag settings 
suggests the improvement is systematic rather than due to 
random variation. While CRN maintains a competitive RMSE, KGCM-VAE 
achieves consistently lower treatment effect estimation 
error, suggesting that its structured latent-variable 
design may more effectively separate temporal dynamics 
from treatment-dependent variation. Notably, the model's 
performance remains stable as the prediction horizon 
increases; this stability against error accumulation 
and temporal drift highlights its utility for modeling 
evolving physical systems where long-term reliability 
is critical.

The ablation results (Figure ~\ref{fig:ablation_unified}) demonstrate that incorporating MMD regularization consistently reduces PEHE across all treatment lags. From a theoretical standpoint, MMD acts as a distribution-matching regularizer that aligns latent representations between treated and control groups. In scientific settings where observational bias and distributional shift are prevalent, such alignment mitigates confounding and improves counterfactual generalization.

The observed improvement in PEHE indicates that enforcing representation invariance enhances the stability and identifiability of treatment effect estimation. Notably, the performance gap slightly widens as treatment lag increases, suggesting that MMD is particularly effective at constraining representation drift caused by the unobserved Arctic seasonal cycle over longer temporal distances.

\subsection{Real-World Case Study}

The real-world Arctic case study provides qualitative validation of the predicted ITEs through a sensitivity analysis of sea ice thickness responses to SSH and velocity perturbations. This evaluation is necessary because ground-truth counterfactual outcomes cannot be directly observed in real-world Arctic systems.

Previous studies have shown that seasonal variations in oceanic conditions, particularly SSH and geostrophic currents, can influence sea ice thickness \cite{wang2021lasting}. Motivated by these findings, we perform a sensitivity analysis using real Arctic observations and estimate treatment effects under two perturbation scenarios. The treated velocity and SSH variables are defined as

\begin{equation}
v_t^{\text{treat}} = (1 + \beta\gamma_t)\tilde{v}_t, \quad
\mathrm{SSH}_t^{\text{treat}} = (1 + \beta\sigma_t)\widetilde{\mathrm{SSH}}_t.\,
\label{eq}
\end{equation}
where $\gamma_t$ and $\sigma_t$ denote the velocity and SSH modulation factors, respectively, and $\beta$ controls the perturbation strength applied to the observed factual values $\tilde{v}_t$ and $\widetilde{\mathrm{SSH}}_t$.

\begin{figure}[H]
    \centering
    % First Image (reduced to 0.48 to fit side-by-side)
    \includegraphics[width=0.48\textwidth]{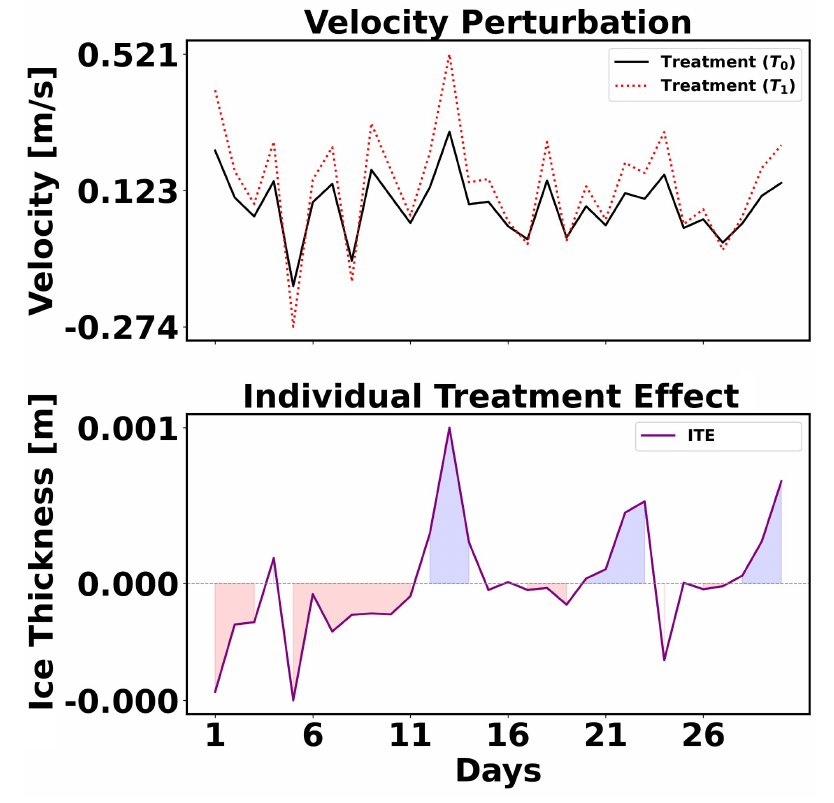}
    \hfill % Adds flexible horizontal space to push them to the outer edges
    % Second Image
    \includegraphics[width=0.48\textwidth]{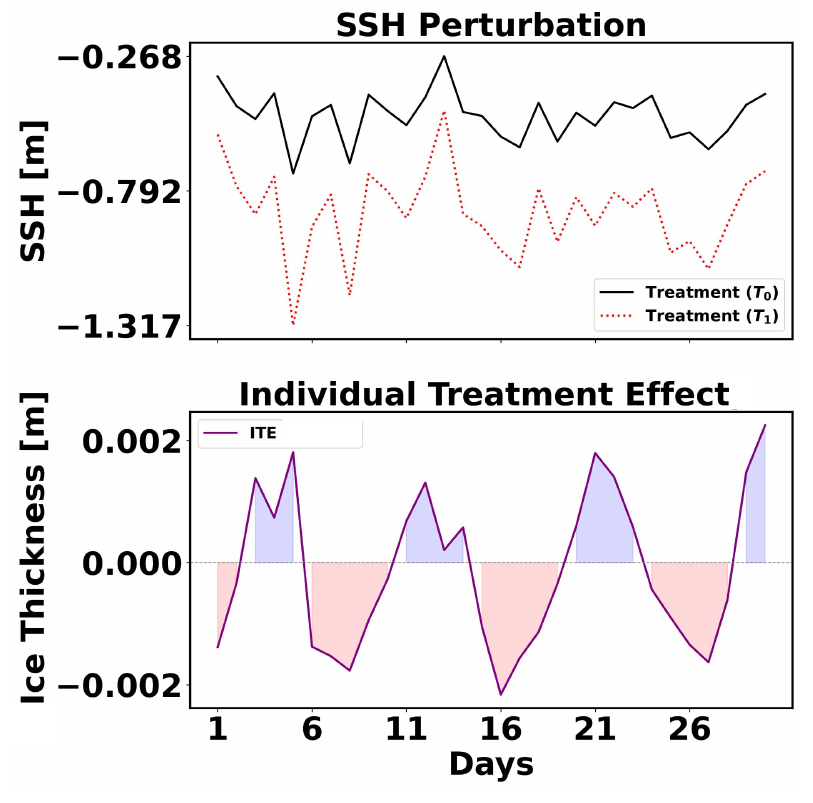}

    \caption{Real-world Individual Treatment Effect (ITE) estimation for a single sequence. (Left) Factual and counterfactual perturbations and predicted ITE in sea ice thickness for velocity perturbations, with blue and red 
shaded regions indicating positive and negative treatment effects, respectively. (Right) Corresponding results for SSH perturbations.}
    \label{fig:arctic_comparison}
\end{figure}
Figure~\ref{fig:arctic_comparison} presents the factual and counterfactual perturbations together with the corresponding ITE estimates. The figure illustrates the temporal evolution of velocity- and SSH-based treatments and the resulting changes in predicted sea ice thickness. Consistent with the modulation scheme in Eq.~\ref{eq}, the KGCM-VAE framework captures distinct non-linear responses to the two forcing mechanisms.

Case Study 1 (Figure~\ref{fig:arctic_comparison}, left panel) examines velocity perturbations. The estimated ITE remains close to zero throughout the 30-day period, with only minor fluctuations around days 13 and 23. This behavior suggests that the imposed velocity perturbation is 
insufficient to drive a substantial change in sea ice thickness, 
as the perturbed velocity remains below the critical threshold 
value $v_0$ defined in the sigmoid modulation scheme 
(Eq.~\ref{eq:sigmoid_modulation}).

Case Study 2 (Figure~\ref{fig:arctic_comparison}, right panel) examines SSH perturbations. In contrast, the estimated ITE exhibits a pronounced oscillatory pattern, alternating between positive and negative values with amplitudes reaching approximately $\pm 0.002$ m. This behavior indicates that SSH-induced forcing generates a stronger response in sea ice thickness than velocity perturbations. The alternating treatment effects directly reflect the physical 
feedback between SSH forcing, ice thinning, and partial recovery, 
a cycle that KGCM-VAE successfully reproduces from data. These findings are consistent with previous studies linking oceanic forcing to sea ice variability \cite{wang2021lasting} and demonstrate that the proposed sigmoid-based treatment formulation can capture complex responses that may be overlooked by linear perturbation approaches.

From a forecasting perspective, the estimated ITEs suggest that SSH anomalies may play a more influential role than surface velocity perturbations in driving short-term sea ice thickness variability. More broadly, these results demonstrate the potential of the KGCM-VAE framework as a tool for causal hypothesis testing in physical systems where controlled experiments are infeasible, enabling researchers to assess the relative influence of competing environmental drivers.
\section{Limitations}

\paragraph{\textbf{Justification of the Adjustment Variable.}}
A key assumption of this work is that total velocity serves as a sufficient proxy to block all confounding paths between SSH and sea ice thickness. We acknowledge that additional atmospheric and oceanic processes, including wind stress, freshwater fluxes, and ocean heat content, may introduce residual confounding that is not fully captured by the velocity field alone. However, total velocity, computed as the magnitude of the eastward and northward velocity components from the ERA5 reanalysis, captures both wind-driven and pressure-driven ocean circulation patterns that govern SSH anomalies and sea ice dynamics in the Arctic. As a result, it represents the most physically comprehensive adjustment variable available within the observational dataset used in this study. The assumption of conditional ignorability given the relevant covariates is standard practice in causal inference with observational data.

\paragraph{\textbf{Causal Graph Specification.}}
The causal graph is constructed from a qualitative literature review rather than data-driven causal discovery. While this ensures physical interpretability, it may not capture all relevant causal pathways present in the observational data. The assumed graph structure therefore represents a simplification of the true Arctic ocean-ice system grounded in known physical relationships.

\section{Conclusion}
Our proposed algorithm, KGCM-VAE, provides a knowledge-guided framework for causal inference in real-world temporal systems by integrating domain knowledge within a variational autoencoder architecture. By explicitly incorporating Maximum Mean Discrepancy (MMD) within the architecture, KGCM-VAE effectively minimizes distributional discrepancies between treated and control covariate distributions in the latent space $\mathbf{z}$. Real-world case studies demonstrate that KGCM-VAE effectively captures the regime-dependent nature of these interactions: velocity perturbations produce negligible effects, consistent with tipping-point theory, whereas SSH perturbations induce oscillatory, non-linear responses as the system repeatedly crosses critical thresholds. This highlights the model’s ability to disentangle treatment-dependent effects from underlying temporal dynamics.

The physics-guided treatment generation, grounded in known relationships between SSH and sea ice thickness, allows KGCM-VAE to learn stable, disentangled representations that capture the underlying confounded system state. As a result, KGCM-VAE achieves consistently lower PEHE compared to established benchmarks. Ablation studies confirm that the inclusion of MMD systematically improves treatment effect estimation, highlighting its effectiveness for causal inference in dynamic, physically constrained systems.

\subsubsection{\ackname} 

This work was supported by the National Science Foundation (NSF) under grants OAC-2118285 (HDR Institute: HARP - Harnessing Data and Model Revolution in the Polar Regions) and OAC-1942714 (CAREER: Big Data Climate Causality Analytics). 

%\subsubsection{\discintname}
%The authors have no competing interests to declare that are relevant to the content of this article.
%\end{credits}
%% Note that this preceding line implies that you store your BibTeX references in a file called 'mybibliography.bib'. If you instead store your references in a file with a different name, for instance 'references.bib', the preceding line should read '\bibliography{references}'. Whatever you do, DO NOT put the file name extension .bib inside the \bibliography command; this will trip up LaTeX compilers. 
%
% If you do not want to use BibTeX, you can also type up the bibliography exactly as you see fit, using the following structure:

\end{document}